\documentclass[letterpaper]{article} 
\usepackage{aaai2026}  
\usepackage{times}  
\usepackage{helvet}  
\usepackage{courier}  
\usepackage[hyphens]{url}  
\usepackage{graphicx} 
\usepackage{natbib}  
\usepackage{caption} 
\usepackage{algorithm}
\usepackage{algorithmic}

\usepackage{amsmath}
\usepackage{cleveref}
\usepackage{booktabs}
\usepackage{multirow}
\usepackage{amssymb}
\crefname{figure}{Fig.}{Figs.}
\Crefname{figure}{Fig.}{Figs.}
\crefname{table}{Tab.}{Tabs.}
\Crefname{table}{Tab.}{Tabs.}
\usepackage{url} 

\usepackage{newfloat}
\usepackage{listings}
\DeclareCaptionStyle{ruled}{labelfont=normalfont,labelsep=colon,strut=off} 
\floatstyle{ruled}
\newfloat{listing}{tb}{lst}{}
\floatname{listing}{Listing}
\nocopyright

\title{Moiré Zero: An Efficient and High-Performance Neural Architecture for Moiré Removal}

\author {
    Seungryong Lee\textsuperscript{\rm 1}\equalcontrib,
    Woojeong Baek\textsuperscript{\rm 2}\equalcontrib,
    Younghyun Kim\textsuperscript{\rm 2},\\
    Eunwoo Kim\textsuperscript{\rm 3},
    Haru Moon\textsuperscript{\rm 3},
    Donggon Yoo\textsuperscript{\rm 3},
    Eunbyung Park\textsuperscript{\rm 2\dag} 
}
\affiliations {
    \textsuperscript{\rm 1}Sungkyunkwan University,
    \textsuperscript{\rm 2}Yonsei University,
    \textsuperscript{\rm 3}Samsung Display
}

\usepackage{bibentry}

\begin{document}

\maketitle

\begin{abstract}
Moiré patterns, caused by frequency aliasing between fine repetitive structures and a camera sensor’s sampling process, have been a significant obstacle in various real-world applications, such as consumer photography and industrial defect inspection. With the advancements in deep learning algorithms, numerous studies---predominantly based on convolutional neural networks---have suggested various solutions to address this issue. 
Despite these efforts, existing approaches still struggle to effectively eliminate artifacts due to the diverse scales, orientations, and color shifts of moiré patterns, primarily because the constrained receptive field of CNN-based architectures limits their ability to capture the complex characteristics of moiré patterns.
In this paper, we propose MZNet, a U-shaped network designed to bring images closer to a ‘Moiré-Zero’ state by effectively removing moiré patterns. It integrates three specialized components: Multi-Scale Dual Attention Block (MSDAB) for extracting and refining multi-scale features, Multi-Shape Large Kernel Convolution Block (MSLKB) for capturing diverse moiré structures, and Feature Fusion-Based Skip Connection for enhancing information flow. Together, these components enhance local texture restoration and large-scale artifact suppression.
Experiments on benchmark datasets demonstrate that MZNet achieves state-of-the-art performance on high-resolution datasets and delivers competitive results on lower-resolution dataset, while maintaining a low computational cost, suggesting that it is an efficient and practical solution for real-world applications.

\begin{links}
    \link{Project Page}{https://sngryonglee.github.io/MoireZero}
\end{links}
\end{abstract}

\section{Introduction}
\begin{figure}[t]
\centering
\includegraphics[width=1\columnwidth]{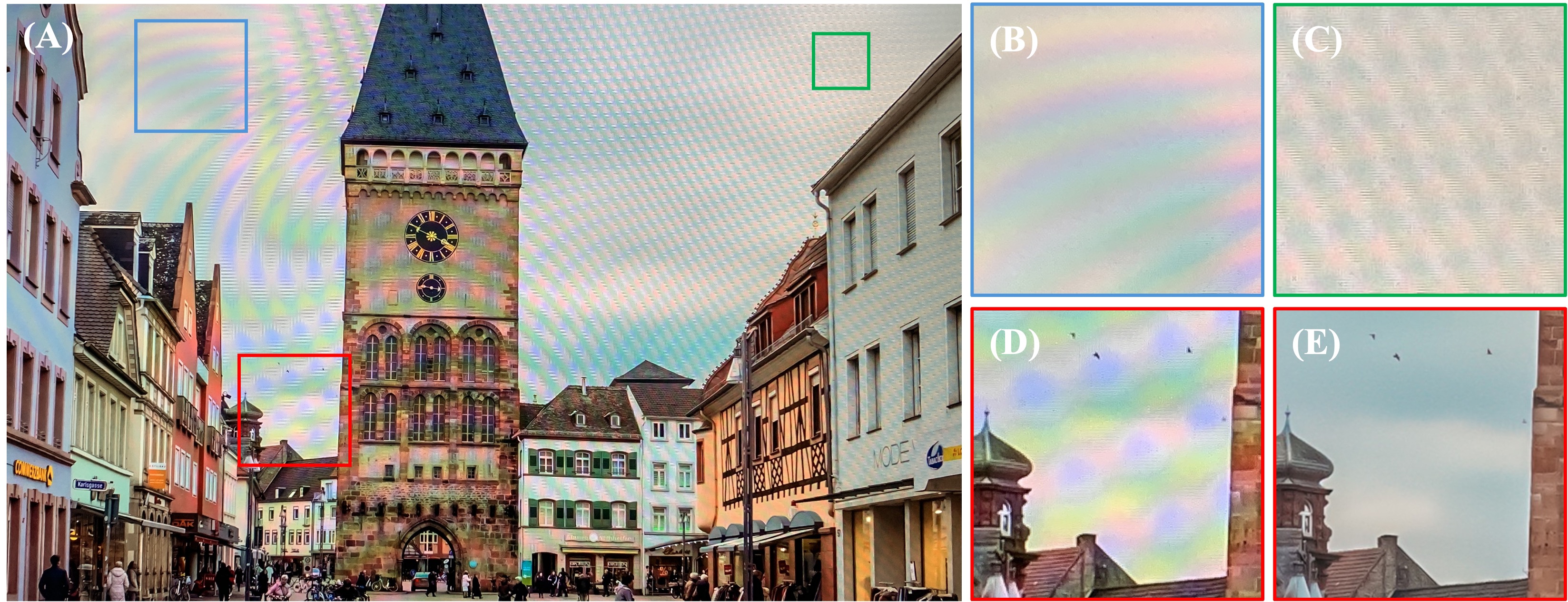}
\caption{Visualization of moir\'e patterns from the UHDM dataset. (A) shows the full image with moiré patterns, while (B), (C), and (D) present magnified regions. They illustrate the varying directions and scales of moiré within a single image. (E) is extracted from the ground truth corresponding to the same region as (D), highlighting the color distortions caused by moiré patterns.}
\label{fig:example}
\end{figure}

Moir\'e patterns arise from frequency aliasing when fine repetitive structures interfere with a camera sensor’s sampling process. These artifacts occur in various scenarios, such as capturing on-screen content, textiles with fine weaves, or architectural structures with dense grids. 
In particular, when photographing digital displays, moir\'e is caused by the interaction between the camera’s color filter array (CFA) and the display’s subpixel structure, resulting in severe color banding and rippling, which significantly degrade image quality (See~\cref{fig:example}).
In industrial applications, such as display manufacturing and quality control, moir\'e artifacts can obscure critical defects during inspection, potentially leading to product failures.
Therefore, developing efficient and robust moir\'e removal techniques has become urgent and essential.

Conventional methods~\cite{liu2015moire, siddiqui2009hardware, sun2014scanned, yang2017textured, yang2017demoireing}, which rely on decomposition, filtering, or handcrafted priors, have attempted to mitigate moir\'e artifacts, but their effectiveness remains limited. With the advent of deep learning, various demoir\'e techniques have emerged, exploring frequency-domain strategies~\cite{zheng2020imagedemoireinglearnablebandpass, zheng2021learning, liu2020waveletbaseddualbranchnetworkimage} and multi-scale architectures~\cite{sun2018moire, cheng2019multiscaledynamicfeatureencoding, yu2022towards}. For example, MBCNN~\cite{zheng2020imagedemoireinglearnablebandpass} employs a learnable bandpass filter to target moir\'e frequencies, while ESDNet~\cite{yu2022towards} incorporates a semantic-aligned scale-aware module for UHD demoir\'eing, achieving promising results.

Despite these advances, moir\'e removal remains challenging due to its diverse scales, complex shapes, and severe color distortions as shown in~\cref{fig:example}. High-definition images further exacerbate this issue, as moir\'e artifacts tend to be more intricate and varied within a single image~\cite{yu2022towards}.
While Transformer-based models have demonstrated promising results in image restoration by capturing long-range spatial dependencies, their high computational cost and the limited availability of large-scale training pose significant challenges for their subsequent application to moir\'e removal. Consequently, almost all existing approaches adopt CNN-based architectures thanks to their efficiency. However, due to their limited receptive fields, CNNs often struggle to capture global structures and unique moir\'e patterns, resulting in suboptimal artifact suppression across diverse scenarios. To comprehensively address these challenges, we propose a novel framework, MZNet, which integrates three complementary components to form a design well-suited to the characteristics of moir\'e patterns.

First, we introduce the Multi-Scale Dual Attention Block (MSDAB), which is designed to effectively capture different scales of the moir\'e patterns. MSDAB in our network extracts multi-scale features from the given input and refines them using Multi-Dilation Convolution Module (MDCM) and Dual Attention Module (DAM), respectively. MDCM consists of depth-wise dilated convolutions, which expand the receptive field of the network with minimal computational cost, while DAM employs Simplified Channel Attention (SCA)~\cite{chen2022simple} and Large Kernel Attention (LKA)~\cite{guo2023visual} to suppress both local textures and large-scale artifacts.
Given that the moir\'e patterns exhibit diverse shapes and orientations, we further incorporate the Multi-Shape Large Kernel Convolution Block (MSLKB) to better capture directional information of the moir\`e patterns. Placed at the bottleneck of our network, MSLKB utilizes depth-wise large-kernel convolutions with square, horizontal, and vertical shapes. This design not only effectively represents the various features with diverse orientations, but also avoids excessive computational cost of the network, providing a tailored solution for subsequent demoiréing.

Finally, we introduce a Feature Fusion-Based Skip Connection (FFSC), which augments the information flow between the encoder and decoder of our network. By aggregating features extracted from all encoder levels and injecting them into each decoder level, this strategy ensures that each decoder leverages rich, multi-scale contextual information, therefore enabling superior detail reconstruction and notable artifact suppression.

Experiments on high-resolution benchmarks (UHDM, FHDMi) demonstrate that MZNet outperforms state-of-the-art demoiréing methods, achieving PSNR gains of over 0.33 dB and 0.67 dB, respectively, while maintaining a low computational cost among competitive approaches.
To validate the practical applicability of MZNet, we trained and applied the model to moir\'e-corrupted display images captured during real-world display manufacturing inspection processes, confirming that it effectively removes moiré patterns while preserving fine details, such as defects that must be retained for inspection. Our contributions are as follows:

\begin{itemize}
    \item We propose MZNet, which incorporates three specialized components for moir\'e removal: Multi-Scale Dual Attention Block, Multi-Shape Large Kernel Convolution Block, and Feature Fusion-Based Skip Connection.

    \item MZNet demonstrates superior demoiré performance on high-resolution benchmark datasets while maintaining a low computational cost.

    \item We validate the practical applicability of MZNet on real-world display inspection images collected from actual manufacturing processes, confirming its effectiveness in industrial applications.
\end{itemize}

\begin{figure*}[h] 
    \centering
    \includegraphics[width=1\textwidth]{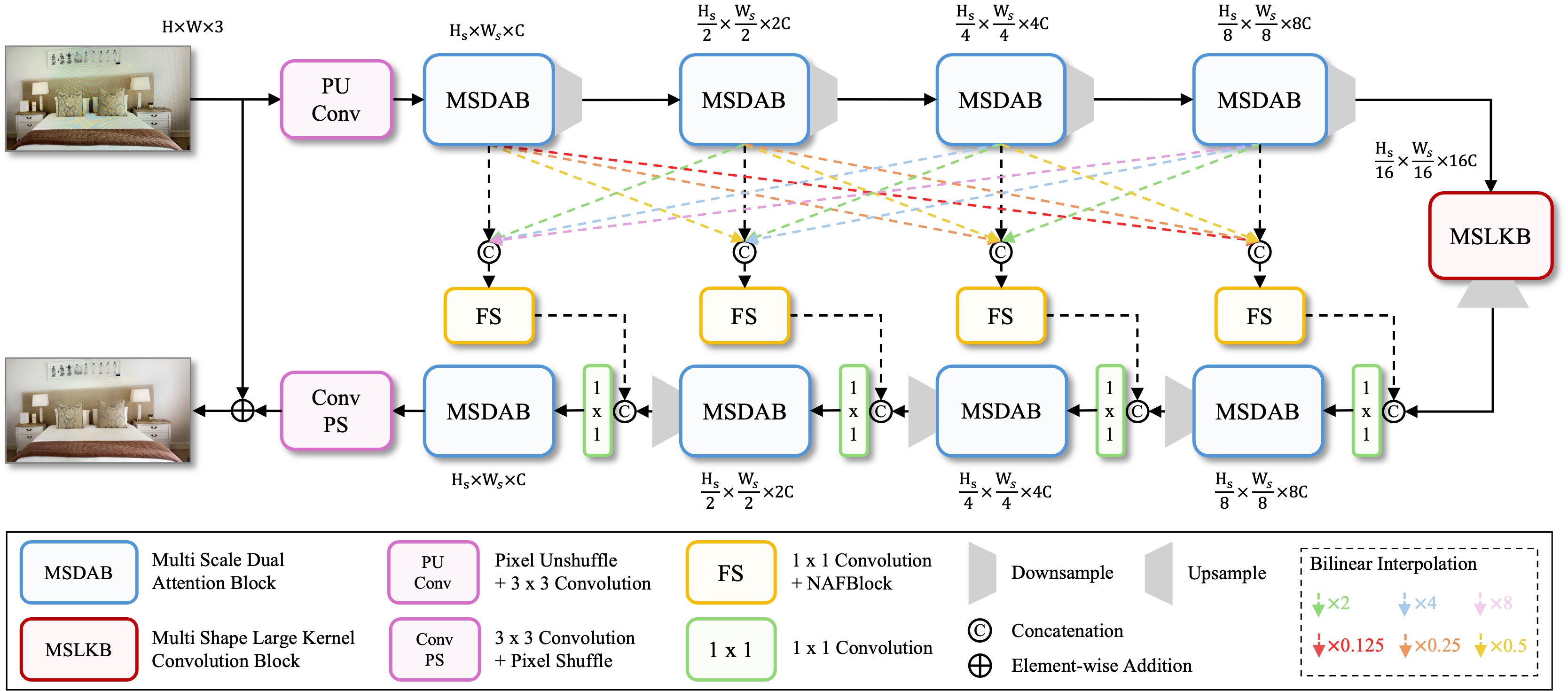}  
    \caption{Overview of MZNet. 
    The model follows a U-Net architecture with four levels of encoders and decoders. MSDAB are used in both the encoder and decoder, while the middle block incorporates a single MSLKB. FS modules in the skip connections facilitate feature fusion across scales.}
    \label{fig:main}
\end{figure*}

\section{Related Work}
\paragraph{Image Demoir\'eing} 
 With the advent of deep learning, various demoir\'e techniques have emerged, exploring frequency-domain strategies~\cite{zheng2020imagedemoireinglearnablebandpass, zheng2021learning, liu2020waveletbaseddualbranchnetworkimage, He2020FHDe2NetFH} and multi-scale architectures~\cite{sun2018moire, he2019mop, cheng2019multiscaledynamicfeatureencoding, yu2022towards}. For example, MBCNN~\cite{zheng2020imagedemoireinglearnablebandpass} employs a learnable bandpass filter to target moir\'e frequencies, while ESDNet~\cite{yu2022towards} incorporates a semantic-aligned scale-aware module for UHD demoir\'eing, achieving promising results. Building on the 4K UHDM benchmark~\cite{yu2022towards}, subsequent studies introduced a coarse‑to‑fine pipeline~\cite{wang2023coarse}, the memory‑efficient patch‑based method\cite{xiao2024pbic}, and RAW‑level approaches that suppress moiré artifacts before demosaicing~\cite{yue2022recaptured, xu2024image, cheng2023recaptured}.
Despite these efforts, existing demoir\'eing methods generally rely on the conventional CNN-based architectures, which often struggle to capture long-range moir\'e structures. 
While many previous approaches have explored multi-scale feature extraction or frequency-based processing, effectively handling the diverse shapes and scales of moiré patterns still poses significant challenges.
In this work, we propose an efficient architecture by enhancing its adaptability to diverse moiré patterns, resulting in a more accurate demoir\'eing network.

\paragraph{Large Kernel Convolution}
Early CNN architectures~\cite{krizhevsky2012imagenet,726791} explored large kernels, but the success of \(3 \times 3\) convolutions in deep neural networks~\cite{simonyan2014very, he2016deep} established them as the standard due to their efficiency. Recently, inspired by Transformer, CNNs have revisited large kernels to model long-range dependencies using depth-wise decompositions~\cite{liu2022more, lau2024large} and large kernel attention mechanisms~\cite{guo2023visual, guo2022segnext}.  
In image restoration, large kernels have been applied for dehazing, deblurring~\cite{luo2023lkd, ruan2023revisiting}, and super-resolution~\cite{wang2024multi}.
In addition, other approaches integrate various kernel structures to jointly capture both global and local information~\cite{cui2024omni, guo2024underwater}.

\paragraph{Image Restoration}
As deep learning advanced, CNN-based models~\cite{anwar2020densely, dudhane2022burst, zamir2020learning, zamir2021multi, zhang2018image, zhang2018residual, gu2019self} achieved strong performance in image restoration.
Following the success of Vision Transformer (ViT)~\cite{dosovitskiy2020image}, Transformer-based models have been widely explored for image restoration, leveraging self-attention to effectively capture global interactions. This approach has been applied to various restoration tasks, including deblurring~\cite{kong2023efficient, tsai2022stripformer, wang2022uformer}, denoising~\cite{chen2020pre, fan2022sunet, li2023efficient}, and super-resolution~\cite{chen2023activating, chen2023dual, liang2021swinir, lu2022transformer, sun2023spatially}. In particular, Restormer~\cite{zamir2022restormerefficienttransformerhighresolution} introduced a transposed self-attention, which applies self-attention across the channel dimension, enabling efficient processing of high-resolution images. However, Transformer-based models often incur high computational costs. To address this issue, NAFNet~\cite{chen2022simple} was proposed, replacing activation functions and Transformer operations with a simplified channel attention and gating mechanism, achieving both high efficiency and strong performance.

\section{Method}\label{sec:method}

\subsection{Overall Pipeline}
As illustrated in~\cref{fig:main}, our framework adopts an encoder-latent-decoder design with four levels on both the encoder and decoder sides. Given an input moir\'e image \(I \in \mathbb{R}^{H \times W \times 3}\), we first apply a pixel unshuffle~\cite{shi2016real} operation followed by a \(3 \times 3\) convolution to extract the shallow feature \(F_0 \in \mathbb{R}^{H_s \times W_s \times C}\). This feature is then passed through the four-stage encoder, which progressively halves the spatial resolution while doubling the number of channels at each level. Consequently, for each level $l$, we obtain latent features \(F_l \in \mathbb{R}^{\frac{H_s}{2^l} \times \frac{W_s}{2^l} \times 2^l C}\). Before decoding, the last latent feature \(F_4\) is refined by our proposed Multi-Shape Large Kernel Convolution Block (MSLKB). 
On the decoder side, the features are upsampled by a factor of 2 at each of the four stages, while the channel dimension is reduced accordingly.
At each decoder level \(l\), the latent features from all encoders are spatially aligned to match the decoder's current resolution \(\frac{H_s}{2^l} \times \frac{W_s}{2^l}\). These aligned features are then fused through a fusion mechanism, resulting in the fused feature \({\tilde{F}}_{l} \in \mathbb{R}^{\frac{H_s}{2^l} \times \frac{W_s}{2^l} \times 2^lC}\). This feature is concatenated with the current decoder feature and passed through a \(1 \times 1\) convolution to reduce the channel dimension by half before proceeding to the next level.
Finally, the decoder’s output features are processed through a \(3 \times 3\) convolution followed by pixel shuffle upsampling to reconstruct the residual image \({R} \in \mathbb{R}^{H \times W \times 3}\). The clean image \({\hat{I}}\) is obtained by adding the residual image \({R}\) to the input moir\'e image \({I}\).

\subsection{Multi-Scale Dual Attention Block}
\begin{figure}[t]    
    \centering     
    \includegraphics[width=1\columnwidth]{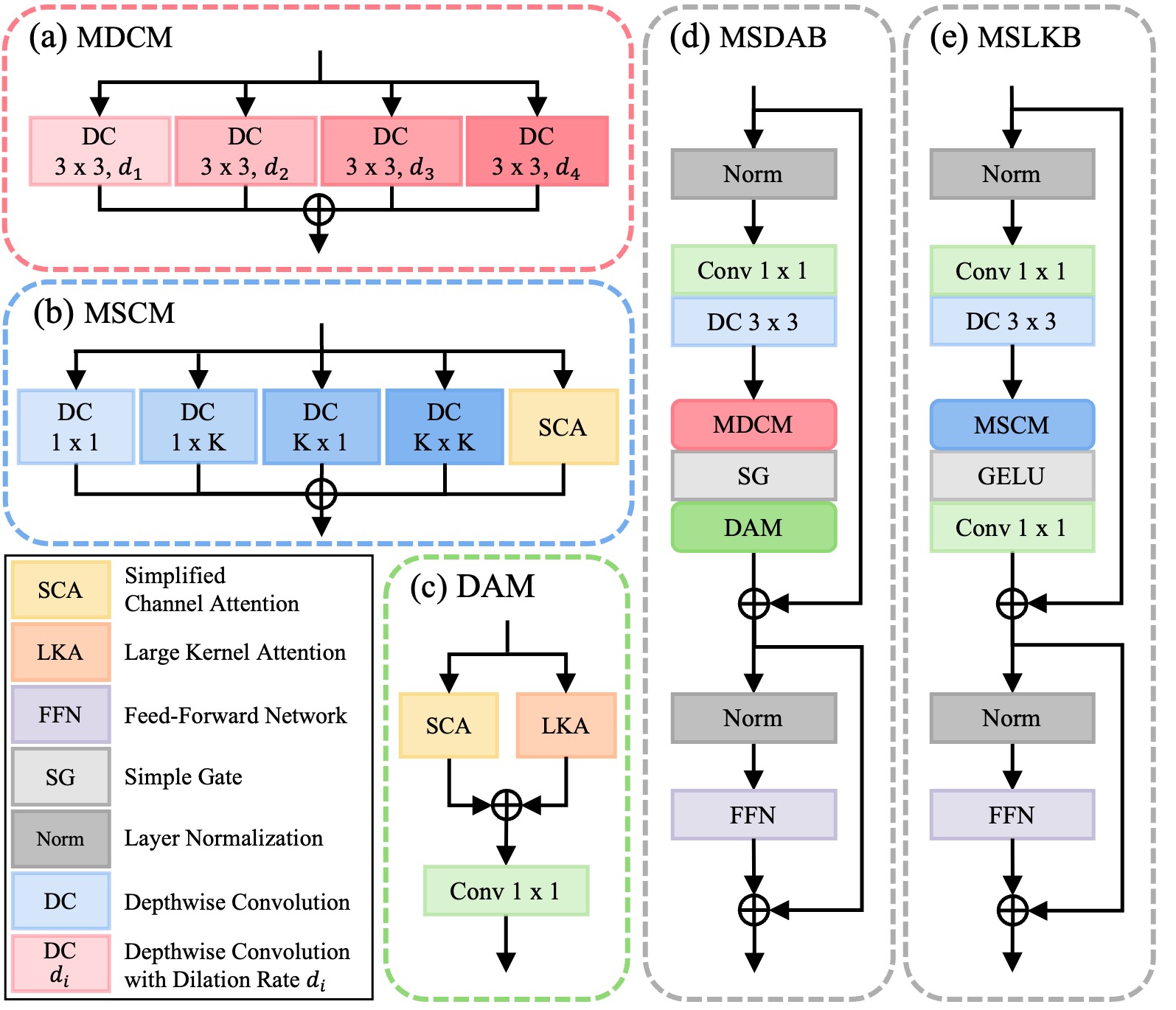} 
    \caption{llustration of the core components in MZNet. Simple Gate from NAFNet serves as an activation function.}
    \label{fig:block}
\end{figure}

To effectively capture and remove moir\'e patterns at varying scales, we propose the Multi-Scale Dual Attention Block (MSDAB), which consists of two key components: the Multi-Dilation Convolution Module (MDCM) and the Dual Attention Module (DAM). MDCM extracts multi-scale features and DAM selectively refines them through complementary attention mechanisms, ensuring effective moir\'e pattern removal.
The input feature map is first normalized and transformed through a \(1\times1\) convolution followed by a \(3\times3\) depth-wise convolution, generating the initial feature representation. 
MDCM then processes this representation by leveraging multiple receptive fields, DAM enhances critical information, and a Feed-Forward Network (FFN) finalizes the output. The structure of MSDAB is illustrated in~\cref{fig:block}d.

\subsubsection{Multi-Dilation Convolution Module}
MDCM is designed to extract multi-scale features by employing multiple parallel \(3 \times 3\) depth-wise convolutions with distinct dilation rates \(\{d_i\}_{i=1}^{4}\), which expand the receptive field to $\{2d_i + 1\}_{i=1}^{4}$.
This design allows the network to capture both fine-grained details and broader spatial structures efficiently. The module is defined as:
\begin{equation}
    \mathrm{MDCM}(X) = \sum_{i=1}^{4} \mathrm{DConv}_{3\times3, d_i}(X),
\end{equation}
where \(X\) is an input feature and \(\mathrm{DConv}_{3\times3, d_i}(\cdot)\) represents a \(3 \times 3\) depth-wise convolution layer with dilation factor \(d_i\).
Inspired by GoogLeNet~\cite{szegedy2015going}, which utilizes parallel convolutional branches, our method adopts a parallel implementation of multiple convolutions with distinct dilation rates to ensure comprehensive coverage of  moir\'e patterns while maintaining computational efficiency through depth-wise convolutions.

\subsubsection{Dual Attention Module}
DAM refines the multi-scale features extracted by MDCM, selectively enhancing important spatial and channel-wise information. It incorporates two complementary attention mechanisms: Large Kernel Attention (LKA)~\cite{guo2023visual} and Simplified Channel Attention (SCA)~\cite{chen2022simple}. LKA captures long-range spatial dependencies through an expanded receptive field, while SCA focuses on the most relevant frequency components by adaptively reweighting feature channels. The attention-enhanced features are computed as:
\begin{equation}
    \mathrm{DAM}({X}) = \mathrm{Conv}_{1\times1}(\mathrm{SCA}({X}) + \mathrm{LKA}({X}) ).
\end{equation}
First, the feature is processed by LKA and SCA modules, and these outputs are fused using a \(1\times1\) convolution to integrate both channel-wise and spatially significant features.
By integrating LKA and SCA, the DAM module ensures that both ``what is important" and ``where it is important" are jointly considered, leading to more precise moir\'e pattern removal.

\subsection{Multi-Shape Large Kernel Convolution Block}
Prior works~\cite{cui2024omni, guo2024underwater} have shown that strip-shaped depth-wise convolutions improve the capture of elongated structures.
However, despite the prevalence of such structures in moiré patterns, research on utilizing different kernel shapes for moiré removal remains underexplored.
Given the varied nature of moiré artifacts, we propose the Multi-Shape Large Kernel Convolution Block (MSLKB) within the latent space of an encoder-decoder architecture to better capture the diverse spatial characteristics of moiré patterns.
MSLKB integrates the Multi-Shape Convolution Module (MSCM) as its core component, applying parallel depth-wise convolutions with diverse kernel forms.
The structures of MSLKB and MSCM are illustrated in~\cref{fig:block}e and ~\cref{fig:block}b, respectively.

Specifically, MSCM employs kernels of size \(K \times K\), \(K \times 1\) and \(1 \times K\) for horizontal and vertical structures, and \(1 \times 1\) to refine local details.
By employing a large K, MSLKB expands the receptive field, enabling the network to capture long-range moir\'e structures more effectively. However, since large kernels tend to incur high computational costs, we place MSLKB at the bottleneck of the network, where the feature resolution is the lowest.
We also adopt SCA in MSCM to selectively enhance the most informative feature channels, complementing the extraction of diverse structural patterns. By integrating these operations in a parallel manner, MSCM is formulated as:

\begin{equation}
    \begin{split}
        \mathrm{MSCM}(X) = \sum_{s \in S} \mathrm{DConv}_{s,1}(X) + \mathrm{SCA}(X), \\
        \quad S = \{K\! \times\! K, K\! \times\! 1, 1\! \times\! K, 1\! \times\! 1\}.
    \end{split}
\end{equation}
By fusing the outputs of these branches, MSCM effectively integrates multi-shape representations, ensuring robust  moir\'e pattern removal. The kernel size $K$ of MSLKB is set to the
largest odd number that does not exceed the size of the bottleneck feature.

\begin{table*}[t]
    \centering
    \small
    \setlength{\tabcolsep}{2pt}    
    \begin{tabular}{c|c|c c c c c c c c c c c| c }
        \toprule
        Dataset & Metrics & Input & DMCNN & MDDM & WDNet & MopNet & MBCNN & FHDe$^2$Net & ESDNet & ESDNet-L & MCFNet & P-BiC & MZNet \\
        \midrule
        \multirow{2}{*}{TIP2018} 
        & PSNR$\uparrow$ & 20.30 & 26.77 & - & 28.08 & 27.75 & 30.03 & 27.78 & 29.81 & 30.11 & 30.13 & \textbf{30.56} & \underline{30.18}\\
        & SSIM$\uparrow$ & 0.738 & 0.871 & - & 0.904 & 0.895 & 0.893 & 0.896 & 0.916 & 0.920 & 0.920 & \textbf{0.925} & \underline{0.921} \\
        \midrule
        \multirow{3}{*}{FHDMi} 
        & PSNR$\uparrow$ & 17.974 & 21.538 & 20.831 & - & 22.756 & 22.309 & 22.930 & 24.500 & 24.882 & 24.823 & \underline{25.450} & \textbf{26.120} \\
        & SSIM$\uparrow$ & 0.7033 & 0.7727 & 0.7343 & - & 0.7958 & 0.8095 & 0.7885 & 0.8351 & 0.8440 & 0.8426 & \underline{0.8473} & \textbf{0.8624} \\
        & LPIPS$\downarrow$ & 0.2837 & 0.2477 & 0.2515 & - & 0.1794 & 0.1980 & 0.1688 & 0.1354 & 0.1301 & \underline{0.1288} & 0.1493 & \textbf{0.1042} \\
        \midrule
        \multirow{3}{*}{UHDM} 
        & PSNR$\uparrow$ & 17.117 & 19.914 & 20.088 & 20.364 & 19.489 & 21.414 & 20.388 & 22.119 & 22.422 & 22.484 & \underline{23.30} & \textbf{23.632} \\
        & SSIM$\uparrow$ & 0.5089 & 0.7575 & 0.7441 & 0.6497 & 0.7572 & 0.7932 & 0.7496 & 0.7956 & 0.7985 & 0.8001 & \underline{0.8007} & \textbf{0.8096} \\
        & LPIPS$\downarrow$ & 0.5314 & 0.3764 & 0.3409 & 0.4882 & 0.3857 & 0.3318 & 0.3519 & 0.2551 & 0.2454 & 0.2536 & \underline{0.2324} & \textbf{0.2237} \\
        \midrule
        \multirow{2}{*}{-} 
        & Params$\downarrow$ (M) & - & \textbf{1.426} & 7.637 & \underline{3.360} & 58.565 & 14.192 & 13.571 & 5.934 & 10.623 & 6.181 & 4.922 & 14.824 \\
        & MACs$\downarrow$ (T) & - & 2.258 & 3.679 & 1.757 & - & 8.522 & 33.23 & 2.247  & 3.689 & 6.903 & \underline{1.223} & \textbf{1.190} \\
        \bottomrule
    \end{tabular}
    \caption{Quantitative comparisons on benchmark datasets. Bold and underlined values indicate the best and second-best results, respectively. Parameters and MACs are for our configuration on the UHDM dataset.}
    \label{tab:performance_comparison}
\end{table*}

\subsection{Feature Fusion-Based Skip Connection}
Instead of one-to-one feature passing scheme used in U-Net~\cite{ronneberger2015u}, our method enables global feature aggregation, enriching decoder representations with multi-scale information.
Let \(F_l\) be the feature map obtained from the \(l\)-th encoder level. To ensure consistency in spatial resolution across all levels, each feature map is resized using bilinear interpolation to match the resolution of the target level, which is given by \(\frac{H_s}{2^l} \times \frac{W_s}{2^l}\). The resized feature maps are then concatenated to form the aggregated feature representation for each level, denoted as \(\bar{F}_l\), which can be written as follows:
\begin{align}
\bar{F}_l &= \mathrm{Concat} ( \{ \mathrm{Interp}(F_k, H_s / 2^l, W_s / 2^l) \}_{k=1}^{4} ),
\end{align}
where \(\mathrm{Interp}(F, h, w)\) represents the bilinear interpolation operation that resizes the feature map $F$ to a spatial resolution of \(h \times w\) and $\mathrm{Concat}(\cdot)$ operator takes a set of interpolated features and concatenates them along the channel dimension.
The aggregated feature map \(\bar{F}_l\) for each layer is then passed through a \(1 \times 1\) convolution to reduce the channel dimension. 
To further refine the aggregated features, we employ Nonlinear Activation-Free Block (NAFBlock)~\cite{chen2022simple}, which improves feature extraction efficiency with a lightweight design.
This results in the final output feature map \(\tilde{F}_l\), expressed as:
\begin{align}
    \ \tilde{F}_{l} = \text{NAFBlock}( \mathrm{Conv}_{1 \times 1}(\bar{F}_l) ).
\end{align} 
By effectively fusing features from all encoder levels, our approach ensures that each decoder level receives rich, multi-scale information. We found that this comprehensive aggregation further helps the decoder to better reconstruct fine details and suppress artifacts.

\subsection{Loss Function}
We adopt a combination of pixel-wise and perceptual losses to train our model. Since moir\'e datasets often contain spatial misalignment between input and ground-truth images (See~\cref{fig:alignment}), the perceptual loss~\cite{johnson2016perceptuallossesrealtimestyle} helps preserve structural details despite these discrepancies.
Following prior works~\cite{zheng2020imagedemoireinglearnablebandpass,yu2022towards}, we supervise the outputs of the final decoder and its two preceding levels. Intermediate predictions are generated via \(3 \times 3\) convolutions and pixel shuffle upsampling. Each level is trained with a combination of pixel-wise and perceptual losses, and the overall loss is defined as:
\begin{equation}
    \mathcal{L} = \mathcal{L}_1 + \lambda \mathcal{L}_\text{lpips}. 
\end{equation}
We empirically set \(\lambda = 1\) for balanced optimization.

\section{Experiments} \label{sec:experiments}
\label{sec:graph}

In this section, we report the qualitative and quantitative results of our proposed MZNet, and conduct ablation studies to validate the efficacy of each component. We also discuss alignment errors in high-resolution Demoiré datasets.

\subsection{Experimental settings}
\paragraph{Datasets and metrics.}
We evaluate our model on three widely used demoir\'eing datasets: TIP2018, FHDMi, and UHDM, and compare its performance with existing state-of-the-art methods. For evaluation, we employ PSNR, SSIM~\cite{1284395}, and LPIPS~\cite{zhang2018unreasonableeffectivenessdeepfeatures}. However, for the TIP2018 dataset, we follow previous works and report only PSNR and SSIM, as LPIPS scores were not previously benchmarked.

\begin{figure*}[t]
    \centering
    \includegraphics[width=1.0\textwidth]{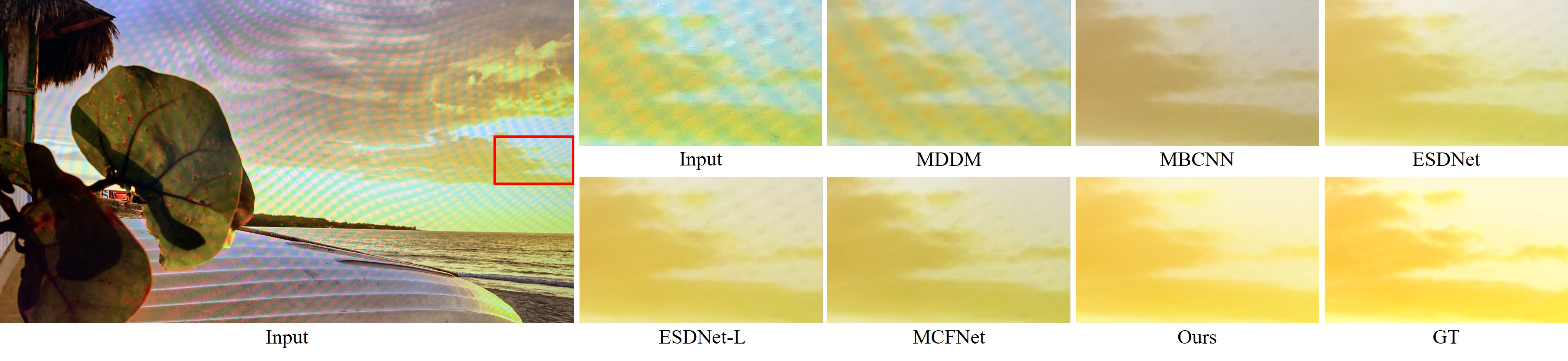}  
    \caption{Qualitative comparisons with state-of-the-art methods on the UHDM dataset.}
    \label{fig:uhdm1}
\end{figure*}


\paragraph{Implementation Details.}
All experiments are conducted on a single NVIDIA GeForce RTX 4090 GPU. Following the training setup of ESDNet~\cite{yu2022towards}, we apply random cropping with a size of 768×768 for the UHDM and 512×512 for the FHDMi. For the TIP2018 dataset, images are first resized to 286×286 and then center-cropped to 256×256. During inference, full resolution images are used. MACs are measured using 4K resolution image.
We optimize the model using Adam optimizer~\cite{kingma2017adammethodstochasticoptimization}, setting \(\beta_1\) and \(\beta_2\) to \(\ 0.9\), and employ a CosineAnnealingLR scheduler~\cite{loshchilov2017sgdrstochasticgradientdescent} to adjust the learning rate throughout training. Detailed training settings for each dataset are included in the supplementary material.

\subsection{Comparison with State-of-the-Art Methods}
\paragraph{Quantitative comparison.}
\cref{tab:performance_comparison} shows that MZNet outperforms prior state-of-the-art methods across all metrics on the high-resolution FHDMi (1920×1080) and UHDM (3840×2160) datasets.
Specifically, on FHDMi, MZNet surpasses P-BiC~\cite{xiao2024pbic} by 0.67 dB in PSNR, and on UHDM, it outperforms P-BiC by 0.33 dB. These results highlight MZNet’s strong ability to handle the complex moiré patterns in high-resolution images.  
On the low-resolution TIP2018 (256x256) dataset, MZNet still achieves the second-best result, demonstrating competitive performance under simpler moiré conditions.  
Notably, despite its high accuracy, MZNet achieves lower computational cost among the compared models, confirming its high accuracy and computational efficiency.

\paragraph{Qualitative comparison.}
\cref{fig:uhdm1} provides a visual comparison on the UHDM datasets, highlighting the qualitative advantages of our method over existing approaches. 
By leveraging large kernels and dilation to expand the receptive field, our model effectively addresses the limitations of existing approaches in removing large-scale moiré patterns. Additional qualitative results in the supplementary material further substantiate its effectiveness.

\begin{table}[t]
\centering
\small
\setlength{\tabcolsep}{4pt} 
\begin{tabular}{c c c c | c c c}
    \toprule
    & \textbf{MSDAB} & \textbf{FFSC} & \textbf{MSLKB} & \textbf{PSNR$\uparrow$} & \textbf{SSIM$\uparrow$} & \textbf{LPIPS$\downarrow$} \\
    \midrule
    (a) &             &              &              & 22.70 & 0.792 & 0.242 \\ 
    (b) & \checkmark  &              &              & 23.04 & 0.797 & 0.240 \\
    (c) &             & \checkmark   &              & 22.86 & 0.791 & 0.226 \\
    (d) &             &              & \checkmark   & 22.88 & 0.792 & 0.249 \\
    (e) & \checkmark  & \checkmark   &              & 23.16 & 0.798 & 0.226 \\
    (f) & \checkmark  &              & \checkmark   & 23.04 & 0.796 & 0.237 \\
    (g) &             & \checkmark   & \checkmark   & 22.97 & 0.795 & 0.238 \\
    (h) & \checkmark  & \checkmark   & \checkmark   & \textbf{23.23} & \textbf{0.801} & \textbf{0.224} \\
    \bottomrule
\end{tabular}
\caption{Ablation studies of each component in MZNet.}
\label{tab:ablation1}
\end{table}

\begin{figure}[t]
    \centering
    \includegraphics[width=0.9\columnwidth]{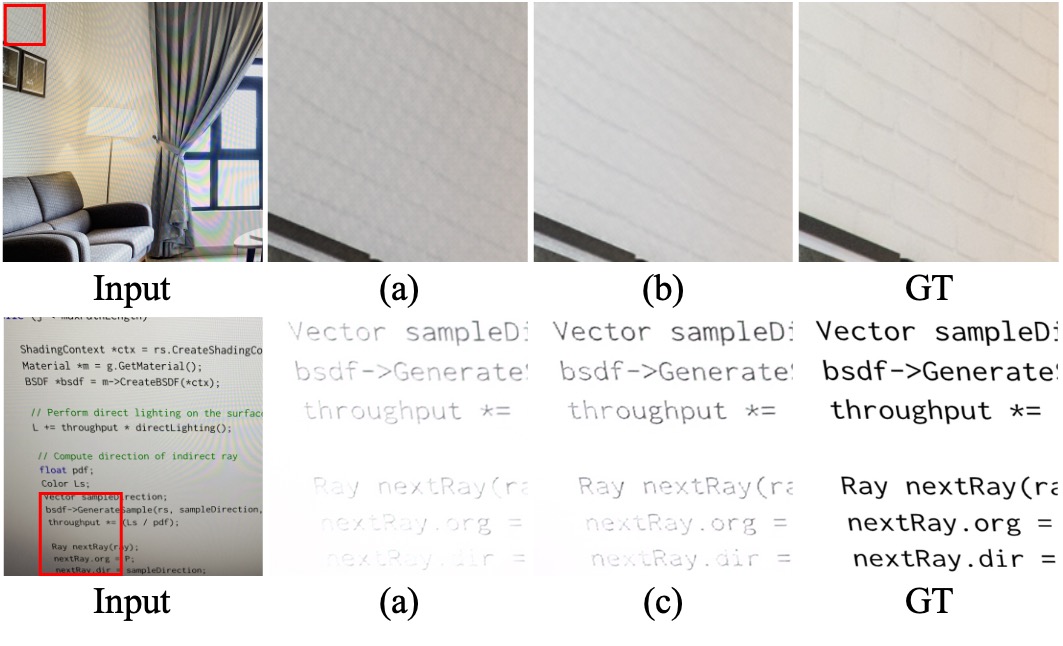}
    \caption{Qualitative comparisons of ablation study.}
    \label{fig:ab_visual}
\end{figure}

\paragraph{Computational Cost Analysis.}
While our model exhibits a relatively large number of parameters due to the use of high channel widths and subsequent \(1\times1\) convolutions, most of the \(3\times3\) convolutions are implemented as depthwise convolution (as shown in~\cref{fig:block}), which helps reduce the overall MACs compared to other methods.
This design balances computational efficiency and feature capacity, ensuring sufficient representation power to handle the complex moiré patterns in high-resolution images while keeping the computational cost low.

\begin{table}[t]
\centering
\small
\setlength{\tabcolsep}{4pt}
\begin{tabular}{l l c c c}
\toprule
\textbf{Block} & \textbf{Method} & PSNR$\uparrow$ & SSIM$\uparrow$ & LPIPS$\downarrow$ \\
\midrule
\multirow{3}{*}{MSDAB} 
    & w/o LKA     & 22.75 & 0.792 & 0.236 \\
    & w/o SCA     & 23.03 & 0.799 & 0.229 \\
    & w/o MDCM    & 23.18 & 0.799 & 0.228 \\
\midrule
\multirow{3}{*}{MSLKB}
    & w/o SCA     & 23.15 & 0.799 & 0.229 \\
    & w/o stripe  & 23.17 & 0.800 & 0.227 \\
    & w/o square  & 23.22 & 0.800 & 0.225 \\
\midrule
\multicolumn{2}{l}{\textbf{Ours}} & \textbf{23.23} & \textbf{0.801} & \textbf{0.224} \\
\bottomrule
\end{tabular}
\caption{Ablation studies on the MSDAB and MSLKB.}
\label{tab:ablation2}
\end{table}

\subsection{Ablation Studies} \label{sec:ablation}
In this section, we analyze the impact of our proposed MZNet by ablating each components of the model. We train a smaller version of our model on the UHDM dataset, using four blocks at each level in both the encoder and decoder.
All models are trained for 200 epochs with a batch size of 2. The initial learning rate is set to 0.0002. All other settings remain consistent with the main experiments.

\paragraph{Effect of Individual Components.}
\cref{tab:ablation1} presents the ablation study results evaluating the contributions of each component in our model. 
The baseline configuration (\cref{tab:ablation1}a) is a simple model where all blocks are NAFBlock~\cite{chen2022simple}, with skip connections implemented by concatenation and a \(1 \times 1\) convolution.
Replacing NAFBlock with MSDAB (\cref{tab:ablation1}b) enhances PSNR to 23.04 dB, demonstrating its effectiveness in handling moiré patterns by capturing multi-scale features with large receptive fields. 
As illustrated in the first row of \cref{fig:ab_visual}, a comparison between (a) and (b) shows that MSDAB better suppresses moiré artifacts.
Incorporating FFSC (\cref{tab:ablation1}c) improves PSNR to 22.86 dB, while MSLKB (\cref{tab:ablation1}d) achieves 22.88 dB, validating their respective contributions to moiré suppression.
Notably, models equipped with FFSC (\cref{tab:ablation1}c, \cref{tab:ablation1}e) significantly improve perceptual quality, reducing LPIPS from 0.242 to 0.226.
This is attributed to FFSC's ability to aggregate multi-scale features across encoder stages, helping the decoder better reconstruct fine details, as seen in the second row of \cref{fig:ab_visual}, where (c) better preserves subtle structures that are lost in (a).
Integrating all components (\cref{tab:ablation1}h) yields the best performance, showing that each module contributes to moiré removal and their combination is most effective. Despite its compact size, our model remains competitive with state-of-the-art methods.

\paragraph{Effect of MSDAB Components}
\cref{tab:ablation2} presents the results of ablating each component of the MSDAB. Removing LKA results in a significant PSNR drop of 0.48 dB, confirming its critical role in the suppressing moir\'e patterns by effectively capturing long-range spatial dependencies. Similarly, removing SCA leads to a PSNR decrease of 0.20 dB, suggesting that channel-wise attention contributes to refining moir\'e-relevant features. Replacing the MDCM with a \(3 \times 3\) depth-wise convolution demonstrates that MDCM is more effective at suppressing moiré distortions compared to a simple convolution.

\paragraph{Effect of MSLKB Components}
\cref{tab:ablation2} presents the ablation study on the key components of the MSLKB. While multiple kernel shapes alone provide a strong baseline, the largest PSNR drop is observed when removing SCA, suggesting its complementary effect to the kernel operations. 
Meanwhile, removing the ``stripe kernel" leads to a slightly greater performance drop than omitting the ``square kernel", suggesting that capturing elongated contextual cues is particularly beneficial for moir\'e removal.
\begin{figure}[t]    
    \centering     
    \includegraphics[width=1\columnwidth]{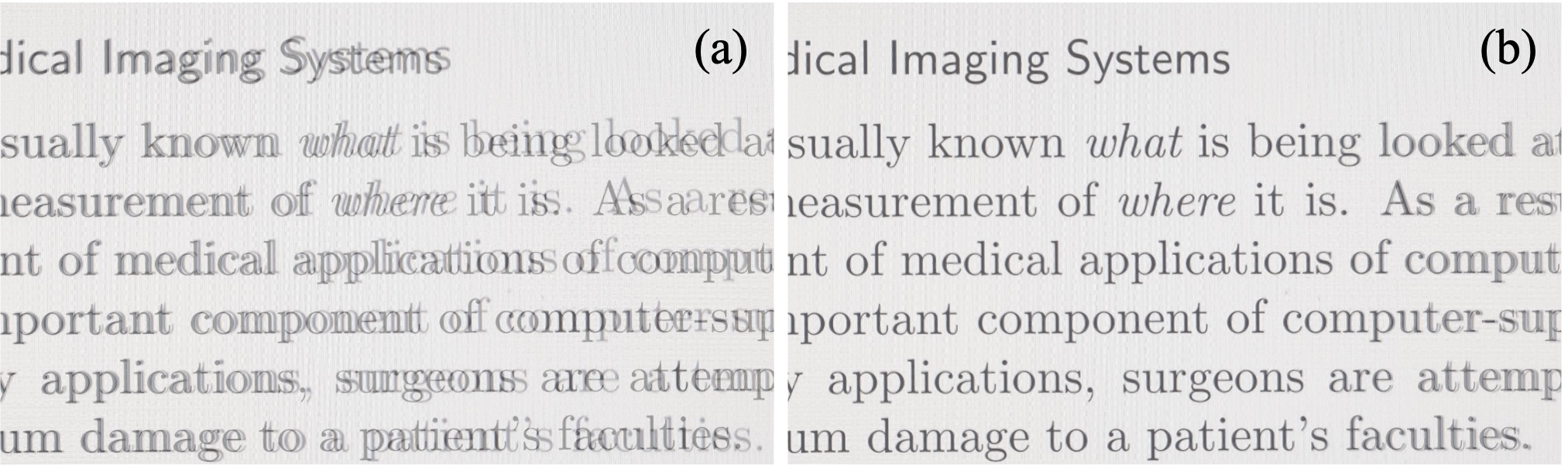} 
    \caption{(a) Overlay of the moiré image and ground truth from the original UHDM dataset, showing noticeable misalignment. (b) Overlay after additional alignment.}
    \label{fig:alignment}
\end{figure}

\begin{table}[t]
\centering
\small
\setlength{\tabcolsep}{3pt} 
    \begin{tabular}{lcccccc}
        \toprule
        \multirow{2}{*}{\textbf{Dataset}} & \multicolumn{3}{c}{\textbf{MZNet (Ours)}} & \multicolumn{3}{c}{\textbf{ESDNet-L}} \\
        \cmidrule(lr){2-4} \cmidrule(lr){5-7}
        & \textbf{PSNR$\uparrow$} & \textbf{SSIM$\uparrow$} & \textbf{LPIPS$\downarrow$} & \textbf{PSNR$\uparrow$} & \textbf{SSIM$\uparrow$} & \textbf{LPIPS$\downarrow$} \\
        \midrule
        Original & 23.63 & 0.810 & 0.223 & 22.42 & 0.799 & 0.245 \\
        Aligned  & \textbf{24.50} & \textbf{0.841} & \textbf{0.205} & \textbf{23.09} & \textbf{0.822} & \textbf{0.235} \\
        \bottomrule
    \end{tabular}
\caption{Impact of UHDM dataset alignment on model performance. Alignment improves all metrics for both models, with a more significant gain in our MZNet.}

\label{tab:alignment_results}
\end{table}
\subsection{Moir\'e Dataset Alignment Issues}
Achieving precise alignment between moiré images and ground truth images remains challenging due to distortions introduced during image capture and inaccuracies in homography estimation, as illustrated in \cref{fig:alignment}. This issue becomes particularly critical for high-resolution datasets such as UHDM, where even slight misalignments can lead to unstable training and significant performance degradation. To mitigate these alignment errors, we further refined the UHDM dataset by applying SIFT-based~\cite{lowe2004distinctive} keypoint detection and matching, leading to improved homography estimation and more precise image warping.

To investigate the impact of dataset alignment issues, we trained MZNet and ESDNet-L on the refined UHDM using identical settings. As shown in \cref{tab:alignment_results}, both models showed substantial improvements across all metrics, underscoring the importance of precise alignment for optimal demoiréing performance. 
These results highlight the need for further research into methods for addressing alignment issues or developing demoiré models more robust to misalignment.

\begin{figure}[h]    
    \centering     
    \includegraphics[width=0.9\columnwidth]{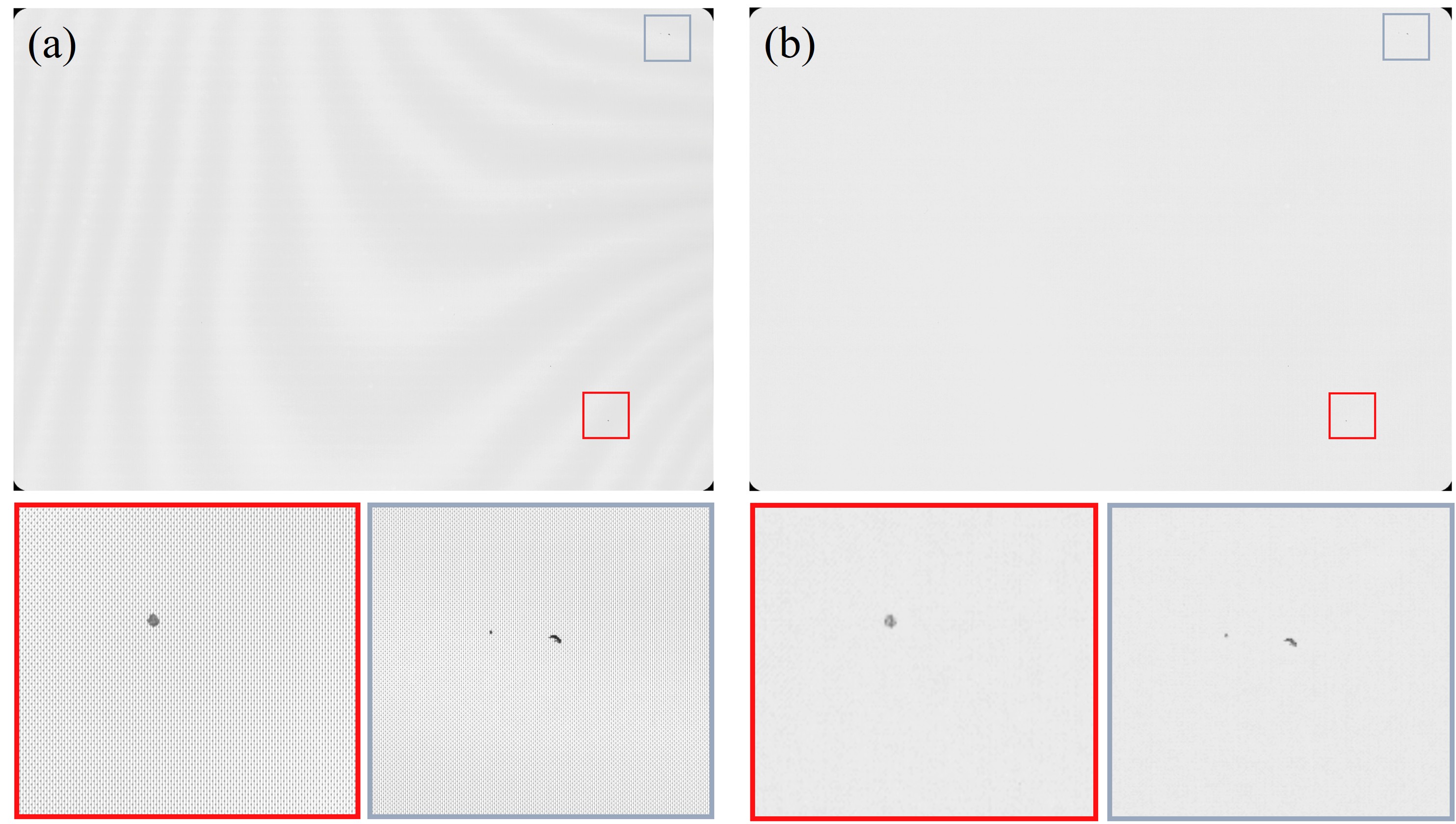} 
    \caption{(a) Input image with moiré, (b) Result of MZNet. An example of removing moir\'e artifacts in display inspection images using MZNet. The zoomed-in section shows that the defects on the display remain intact.}
    \label{fig:sdc}
\end{figure}

\section{Application to Real World Display Inspection}

To evaluate the practical applicability of our model, we deployed it in a display inspection scenario using real-world images from actual manufacturing processes.
In this industrial inspection process, solid-color patterns are typically shown on the display panels, and cameras capture the display to reveal subtle defects such as dead pixels or line irregularities.
However, as shown in~\cref{fig:sdc}a, moiré patterns often appear in the images, obstructing accurate defect detection. 
To mitigate this issue, current industry practices involve defocusing the camera during image acquisition to reduce moiré interference. While this method suppresses moiré, it also reduces image clarity, making precise defect detection more challenging. 
To address this limitation, we trained MZNet on images captured during the display inspection process. As shown in~\cref{fig:sdc}b, our model effectively removes moiré artifacts while preserving fine details, such as defects, with minimal information loss. This enhancement improves the efficiency and reliability of display inspection, eliminating the need for defocusing while maintaining high image quality for accurate defect detection.
We further validated its robustness on inspection images using random noise patterns instead of solid colors (see supplementary material).

\section{Conclusion}
In this study, we propose MZNet, an efficient and effective solution for moiré pattern removal.
Our model integrates MSDAB for multi-scale feature extraction, MSLKB for capturing diverse moiré structures, and FFSC for enhancing information flow, enabling robust moiré elimination while preserving fine details.
Experiments on benchmark datasets demonstrate that MZNet outperforms existing methods in both quantitative metrics and visual quality, achieving state-of-the-art performance in challenging high-resolution scenarios.
We also discuss the impact of moiré dataset alignment issues and demonstrate how precise alignment improves model performance. Its practical applicability has been validated on real-world industrial data, confirming its effectiveness beyond controlled test conditions.

\bibliography{aaai2026}
\clearpage
\appendix
\section{Dataset and Experimental Setup}
We report the detailed hyperparameter configuration of MZNet and dataset statistics. In all experiments, we used a total of 4, 4, 6, and 8 MSDAB blocks in the first to fourth encoder layers, respectively, and 4, 4, 6, and 6 blocks in the decoder. A single Multi-Shape Large Kernel Convolution Block (MSLKB) is employed at the bottleneck layer. The kernel size $K$ of MSLKB is set to the largest odd number that does not exceed the size of the bottleneck feature. For the MDCM module, we empirically found that setting dilation $\{d_i\}_{i=1}^{4}$ to $\{1, 4, 7, 9\}$ yields optimal results, effectively capturing moir\'e patterns at different scales. 
Since our model is trained on cropped patches, we employed the Test-time Local Converter (TLC)~\cite{chu2022improving} to ensure consistency by adapting local features during inference, thereby reducing the discrepancy between the training and inference data. We utilized ptflops~\cite{ptflops} library to measure the MACs of each model.

The number of parameters and MACs for MZNet on each dataset is summarized in Table~\ref{tab:macs_params}. Note that all MACs are measured using 4K resolution inputs to ensure fair comparison across datasets.

\paragraph{TIP2018}
The TIP2018 dataset~\cite{sun2018moire} includes a total of 135,000 image pairs for training and testing. It was created by capturing photographs of ImageNet images displayed on various hardware screens, leading to a diverse range of resolutions and moir\'e patterns. For training, we limited the dilation rates of the MSDAB at the final encoder and decoder levels to $\{1, 4, 7\}$ due to the low-resolution input. we set the kernel size \(K\) of MSLKB to 7, trained for 100 epochs with a batch size of 8, and used an initial learning rate of \(\mathrm{2e^{-4}}\), which decayed to \(\mathrm{1e^{-6}}\) using cosine annealing scheduler.\\

\paragraph{FHDMi}
The FHDMi dataset~\cite{He2020FHDe2NetFH} includes 9,981 training pairs and 2,019 testing pairs, all at a resolution of 1920×1080. The ground truth images were collected based on 18 categories of frequently observed screen content, including wallpapers, sports video frames, film clips, and documents. For training, we set \(K\) of MSLKB to 15, trained for 400 epochs with a batch size of 4, and used an initial learning rate of \(\mathrm{4e^{-4}}\), which decayed to \(\mathrm{1e^{-6}}\) using cosine annealing scheduler.\\

\paragraph{UHDM}
The UHDM dataset~\cite{yu2022towards} contains 5,000 ultra-high-definition image pairs, with 4,500 used for training and 500 for testing. The image resolutions range from 4032×3024 to 4624×3472, covering diverse scenes such as landscapes, sports, videos, and documents. It was collected using multiple mobile devices and display screens to ensure a wide variety of real-world UHD  moir\'e patterns. For training, we set \(K\) of MSLKB to 23, trained for 700 epochs with a batch size of 8 (using gradient accumulation with 4 steps), and used an initial learning rate of \(\mathrm{6e^{-4}}\), which decayed to \(\mathrm{1e^{-6}}\) using cosine annealing scheduler.

\begin{table}[h]
\centering
    \begin{tabular}{ c c | c c c }
        \toprule
        & Metric & \textbf{UHDM} & \textbf{FHDMi} & \textbf{TIP2018} \\
        \midrule
        &  Params$\downarrow$ (M) & 14.82 & 14.66 & 14.49 \\
        & MACs$\downarrow$ (T) & 1.190 & 1.189 & 1.186 \\
        \bottomrule
    \end{tabular}
    \caption{Comparison of model parameters and MACs across datasets. All MACs are measured using 4K-resolution inputs. Minor differences arise due to dataset-specific hyperparameter configurations, such as the kernel size \( K \) of the MSLKB module.}
    \label{tab:macs_params}
\end{table}
\vspace{1cm}



\begin{figure}[h] 
    \centering
    \includegraphics[width=1\columnwidth]{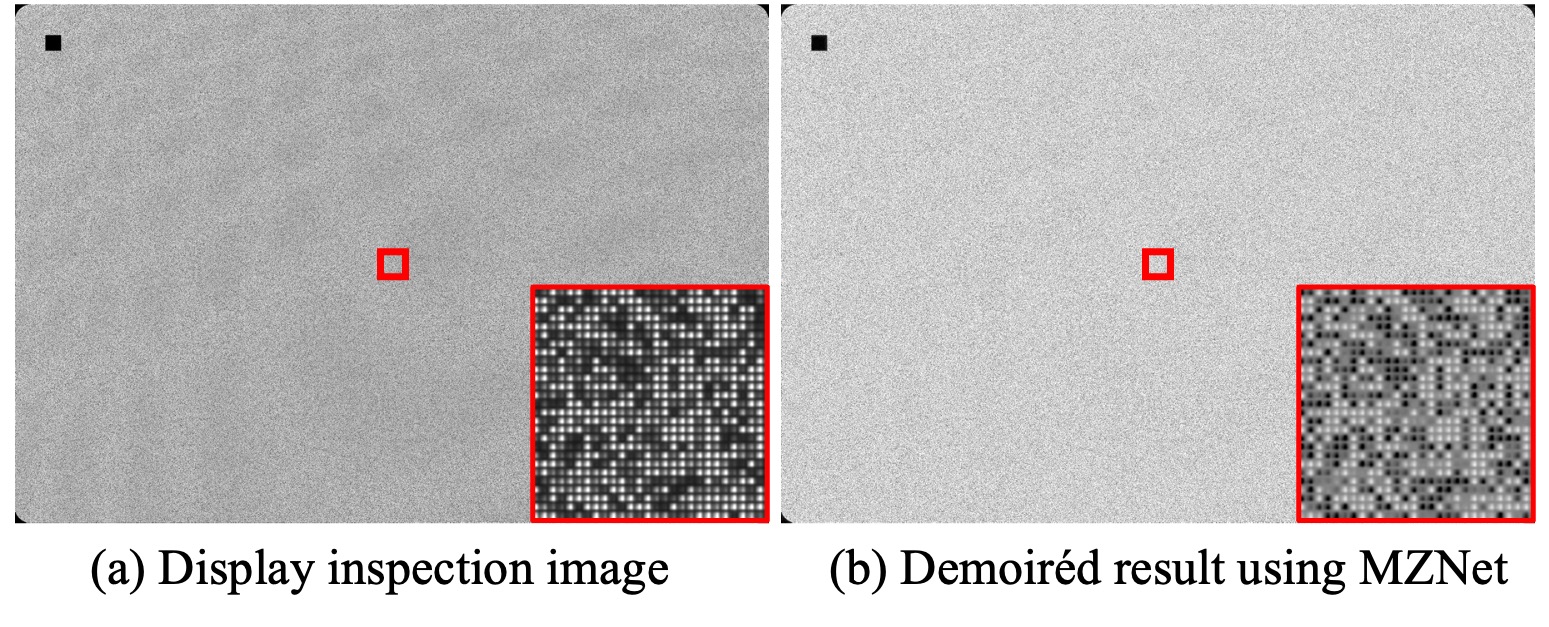}  
    \caption{Challenging inspection scenario with synthetic noise, highlighting the difficulty of moiré removal while preserving pixel integrity.}
    \label{fig:supple_sdc1}
\end{figure}

\begin{table*}[t]
\centering 
\small
\setlength{\tabcolsep}{5pt} 
\begin{tabular}{c |c c| c c c | c c c }
\toprule
\multirow{2.7}{*}{Model} & \multirow{2.7}{*}{Params$\downarrow$ (M)} & \multirow{2.7}{*}{MACs$\downarrow$ (T)} & \multicolumn{3}{c|}{FHDMi} & \multicolumn{3}{c}{UHDM} \\
\cmidrule{4-9}
 &  &  & PSNR $\uparrow$ & SSIM $\uparrow$ & LPIPS $\downarrow$ & PSNR $\uparrow$ & SSIM $\uparrow$ & LPIPS $\downarrow$ \\
\midrule
MBCNN\cite{zheng2020imagedemoireinglearnablebandpass} & 14.192 & 8.522 & 22.31 & 0.8095 & 0.1980 & 21.41 & 0.7932 & 0.3318 \\
FHDe$^2$Net\cite{He2020FHDe2NetFH}  & 13.571 & 33.23 & 22.93 & 0.7885 & 0.1688 & 20.39 & 0.7496 & 0.3519 \\
ESDNet\cite{yu2022towards}  & 5.934 & 2.247 & 24.50 & 0.8351 & 0.1354 & 22.12 & 0.7956 & 0.2551 \\
ESDNet-L\cite{yu2022towards}  & 10.623 & 3.689 & 24.88 & 0.8440 & 0.1301 & 22.422 & 0.7985 & 0.2454 \\
MCFNet\cite{nguyen2023multiscale}  & 6.181 & 6.903 & 24.82 & 0.8426 & 0.1288 & 22.48 & 0.8001 & 0.2536 \\
P-BiC\cite{xiao2024pbic}  & 4.922 & 1.223 & 25.45 & 0.8473 & 0.1493 & 23.30 & 0.8007 & 0.2324 \\
MZNet-Small  & 7.46 & 0.596 & 25.61 & 0.8522 & 0.1111 & 23.26 & 0.8022 & 0.2242 \\
MZNet-Medium & 12.61 & 0.823 & 25.91 & 0.8582 & 0.1078 & 23.47 & 0.8043 & 0.2230 \\
MZNet-Large  & 14.82 & 1.190 & 26.12 & 0.8624 & 0.1042 & 23.63 & 0.8096 & 0.2237 \\
\bottomrule
\end{tabular}
\caption{Performance of lightweight MZNet variants on high-resolution datasets. The “Large” model is the original version reported in the main paper, while “Small” and “Medium” are lightweight variants for supplementary analysis. All models are evaluated using PSNR, SSIM, and LPIPS. Params and MACs are measured on 4k input images.}
\label{tab:lightweight_variants}
\end{table*}
\section{Display Inspection Dataset}

In typical inspection settings, display panels are illuminated with a uniform solid color background. 
The image used in our experiment is ultra-high-resolution (about 7000×5000 pixels).
Additionally, to assess the robustness of our method under more challenging conditions, we conducted supplementary experiments by adding synthetic random noise patterns to the solid color background.  
This setup leads to complex variations in pixel intensity, posing a more challenge for moiré removal. The model must eliminate moiré artifacts while preserving the local intensity contrast and maintaining fine structural details without distortion.
As shown in the red box in~\cref{fig:supple_sdc1}, we provide a magnified region where individual display pixels and their subtle brightness variations are clearly visible, emphasizing the difficulty of the task.

\begin{figure*}[h] 
    \centering
    \includegraphics[width=0.95\linewidth]{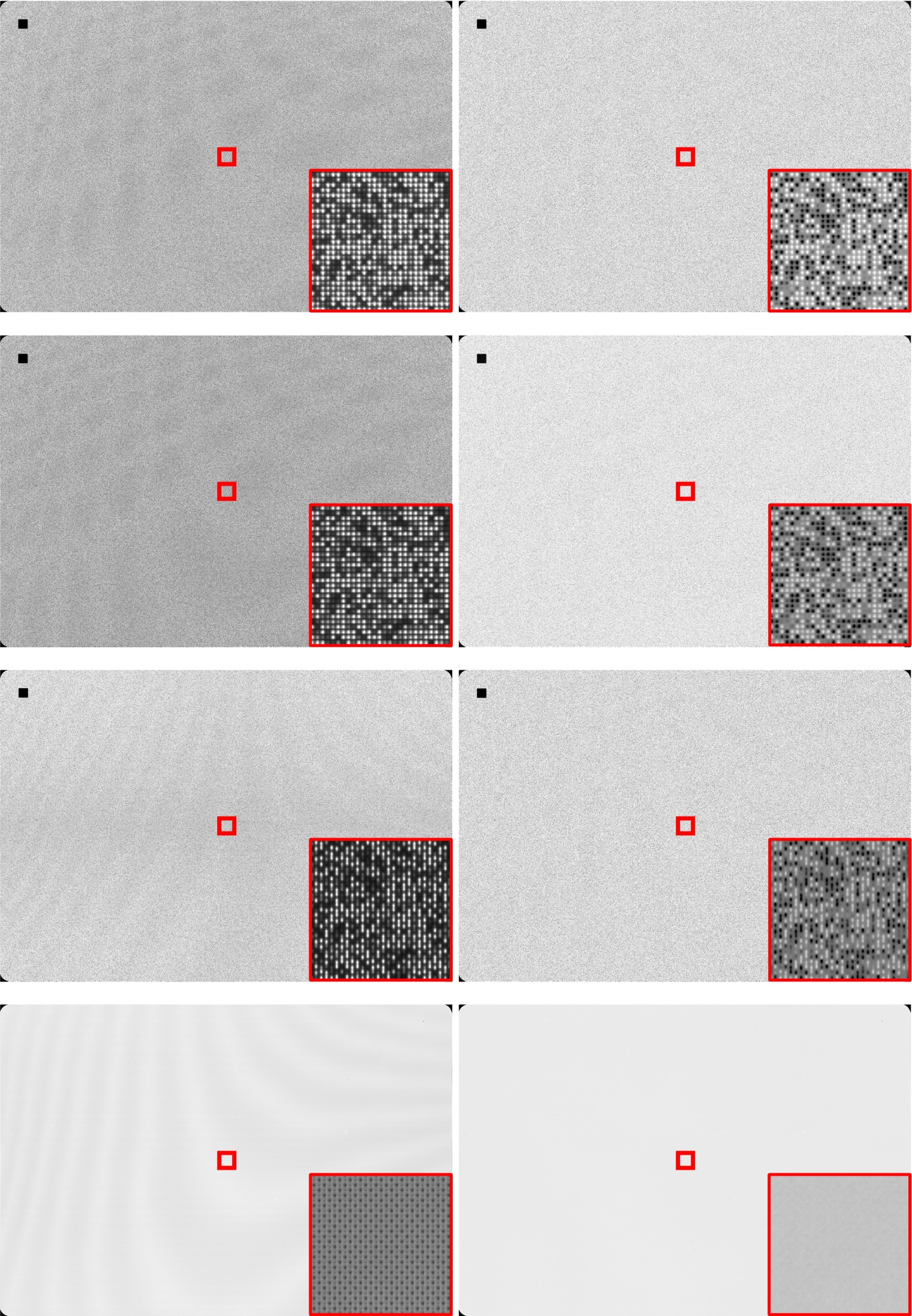}  
    \caption{Examples of moiré artifact removal in display inspection images using MZNet. The left image shows the input with visible moiré patterns, while the right image presents the result after moiré removal.}
    \label{fig:supple_sdc}
    
\end{figure*}

As demonstrated in~\cref{fig:supple_sdc1}, our method effectively removes moiré artifacts while faithfully preserving the pixel-wise brightness differences induced by the display pattern and noise.
Importantly, while absolute brightness values may slightly change after demoiréing, what matters most in industrial inspection is preserving the relative brightness variations between adjacent pixels—particularly those induced by the noise pattern.
These results confirm the practical applicability of MZNet in real-world industrial demoiréing scenarios, where both artifact removal and accurate intensity preservation are essential.
Additional visual examples are provided in~\cref{fig:supple_sdc}

\vspace{1cm}
\section{Evaluation of Lightweight Variants on High-Resolution Datasets}
Although our main model already demonstrates strong performance across all datasets, we additionally evaluate two lightweight variants to verify that the performance gains do not stem solely from the model's parameter count.
Specifically, we design MZNet-Medium and MZNet-Small by reducing the number of MSDAB in each encoder and decoder stage to (2, 2, 4, 6) and (2, 2, 2, 2), respectively. All Multi-Dilation Convolution Module (MDCM) use dilation rates of 1, 4, and 9, and all other experimental settings are kept identical to those of the main model.

These variants are evaluated on high-resolution datasets (FHDMi and UHDM), where moiré artifacts tend to be more complex and the performance gap between different model capacities becomes more apparent.

As shown in~\cref{tab:lightweight_variants}, both lightweight variants show reasonable performance across all metrics, suggesting that the proposed architecture can remain effective even under constrained model size.

\section{More Visual Results}
\subsection{High-resolution Benchmark Datasets}
\begin{figure*}[h] 
    \centering
    \includegraphics[width=1\linewidth]{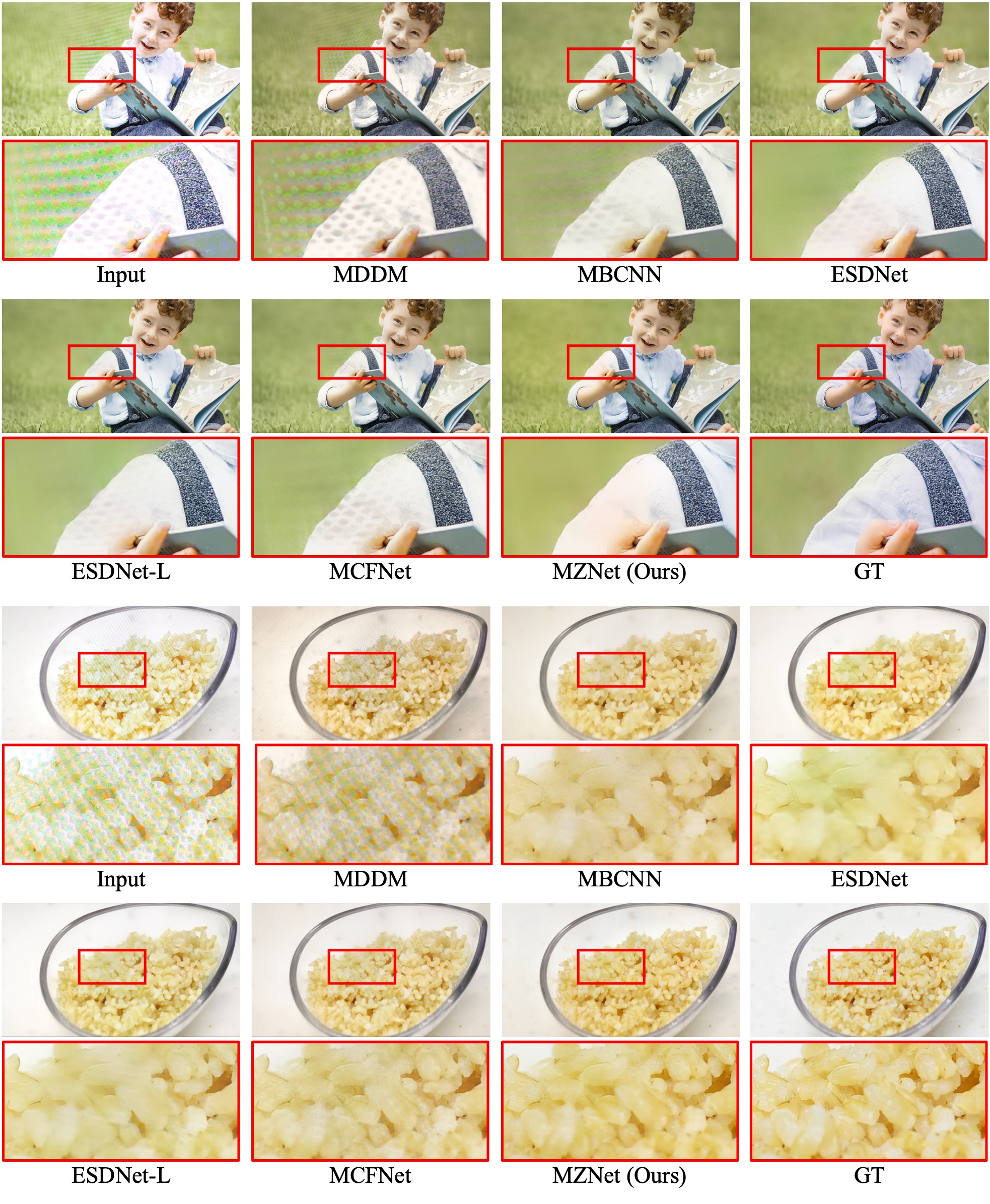}  
    \caption{Qualitative comparisons with state-of-the-art methods on the UHDM dataset.}
    \label{fig:supple_uhdm}
\end{figure*}

\begin{figure*}[h] 
    \centering
    \includegraphics[width=1\linewidth]{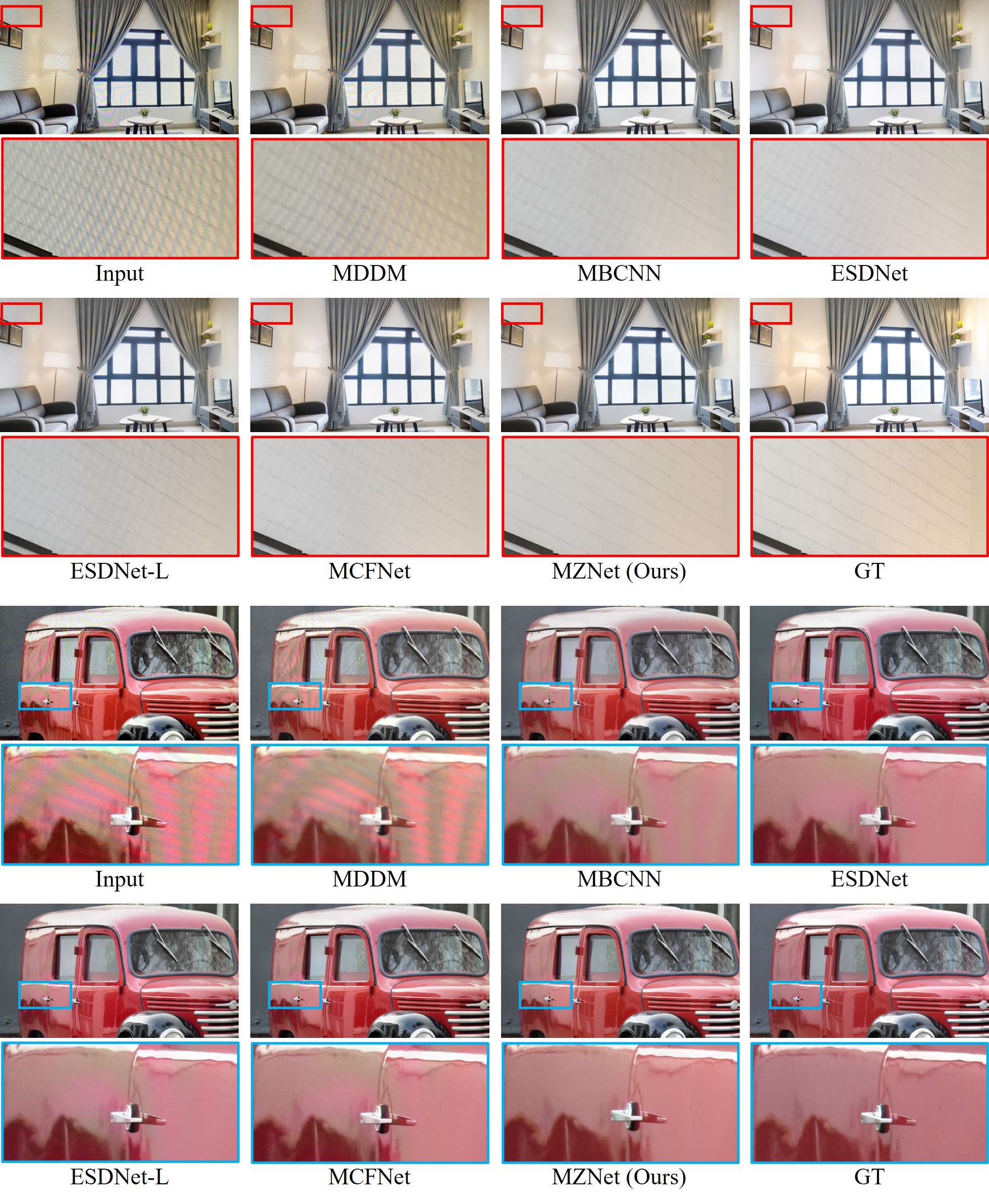}  
    \caption{Qualitative comparisons with state-of-the-art methods on the UHDM dataset.}
    \label{fig:supple_uhdm_2}
\end{figure*}

\begin{figure*}[h] 
    \centering
    \includegraphics[width=1\linewidth]{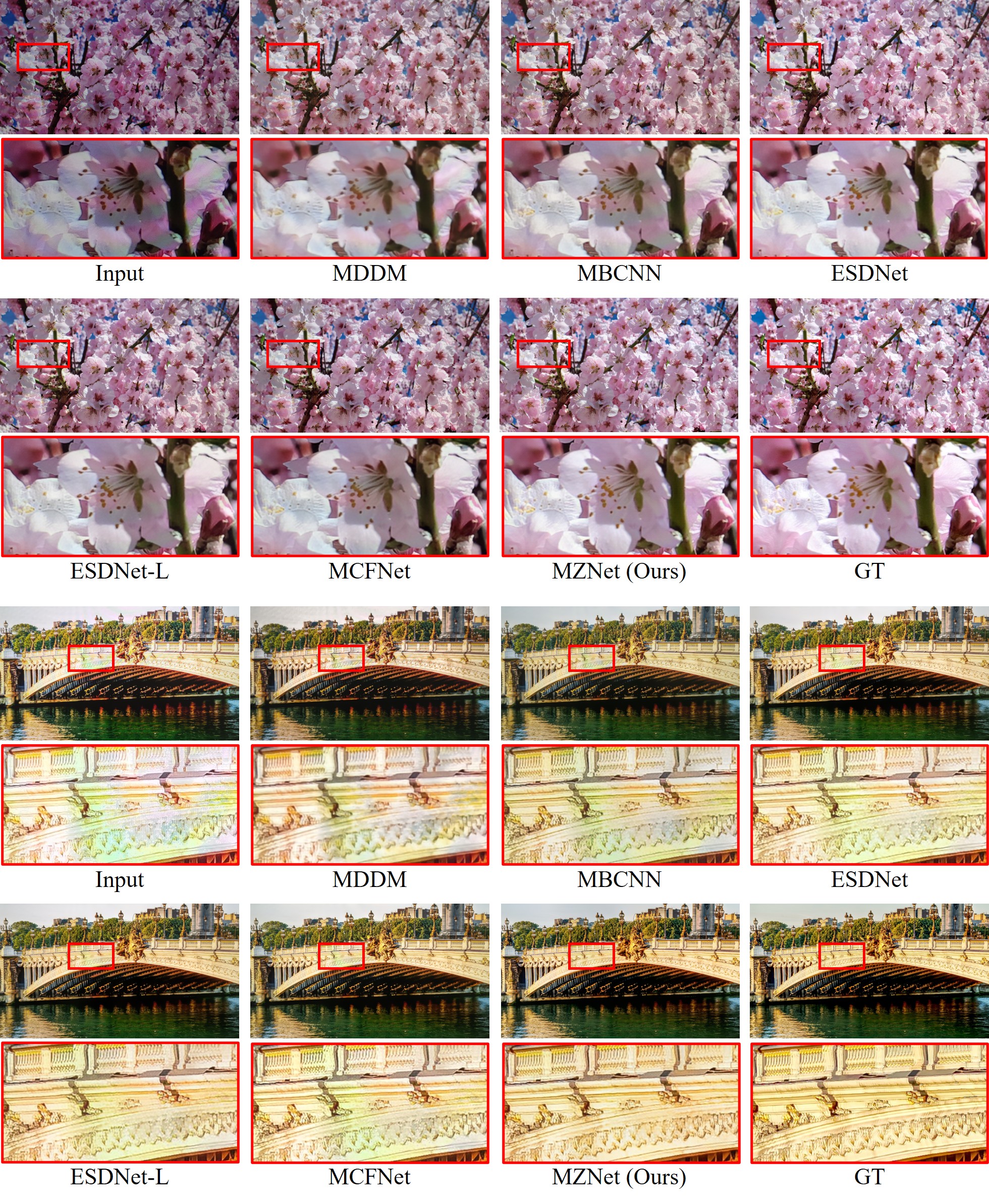}  
    \caption{Qualitative comparisons with state-of-the-art methods on the UHDM dataset.}
    \label{fig:supple_uhdm_3}
\end{figure*}

\begin{figure*}[h] 
    \centering
    \includegraphics[width=1\linewidth]{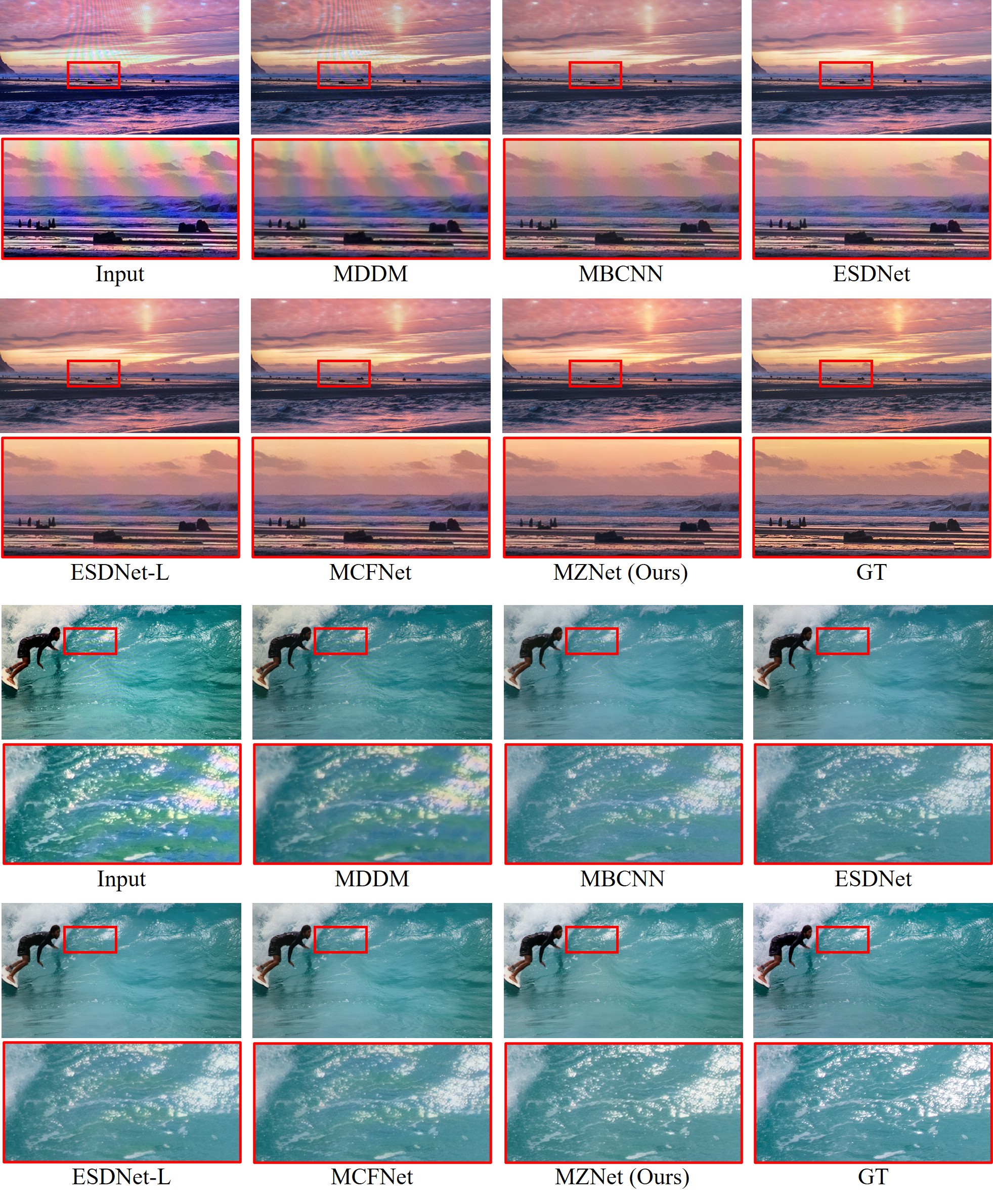}  
    \caption{Qualitative comparisons with state-of-the-art methods on the UHDM dataset.}
    \label{fig:supple_uhdm_4}
\end{figure*}

\begin{figure*}[h] 
    \centering
    \includegraphics[width=1\linewidth]{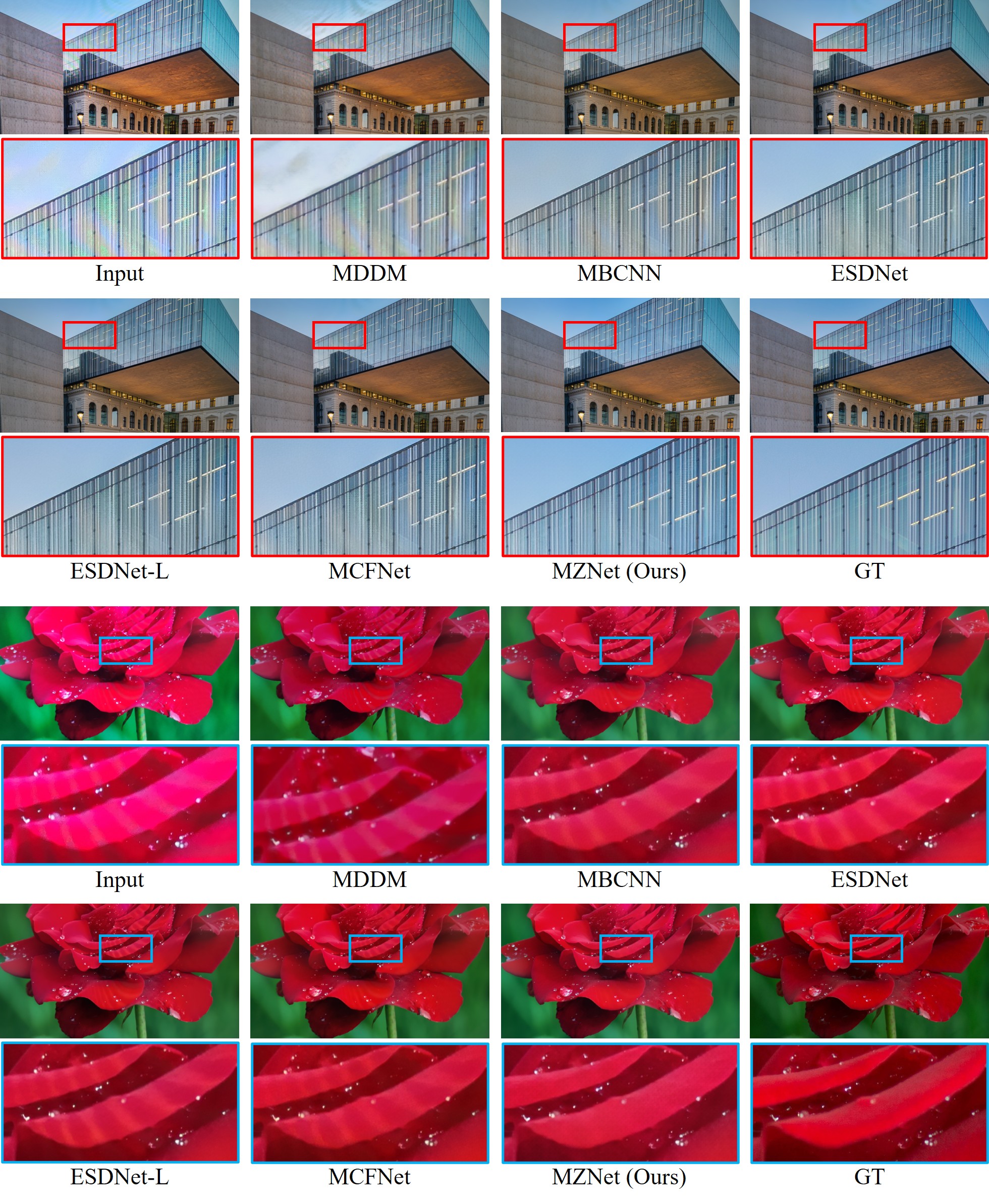}  
    \caption{Qualitative comparisons with state-of-the-art methods on the UHDM dataset.}
    \label{fig:supple_uhdm_5}
\end{figure*}

\begin{figure*}[h] 
    \centering
    \includegraphics[width=1\linewidth]{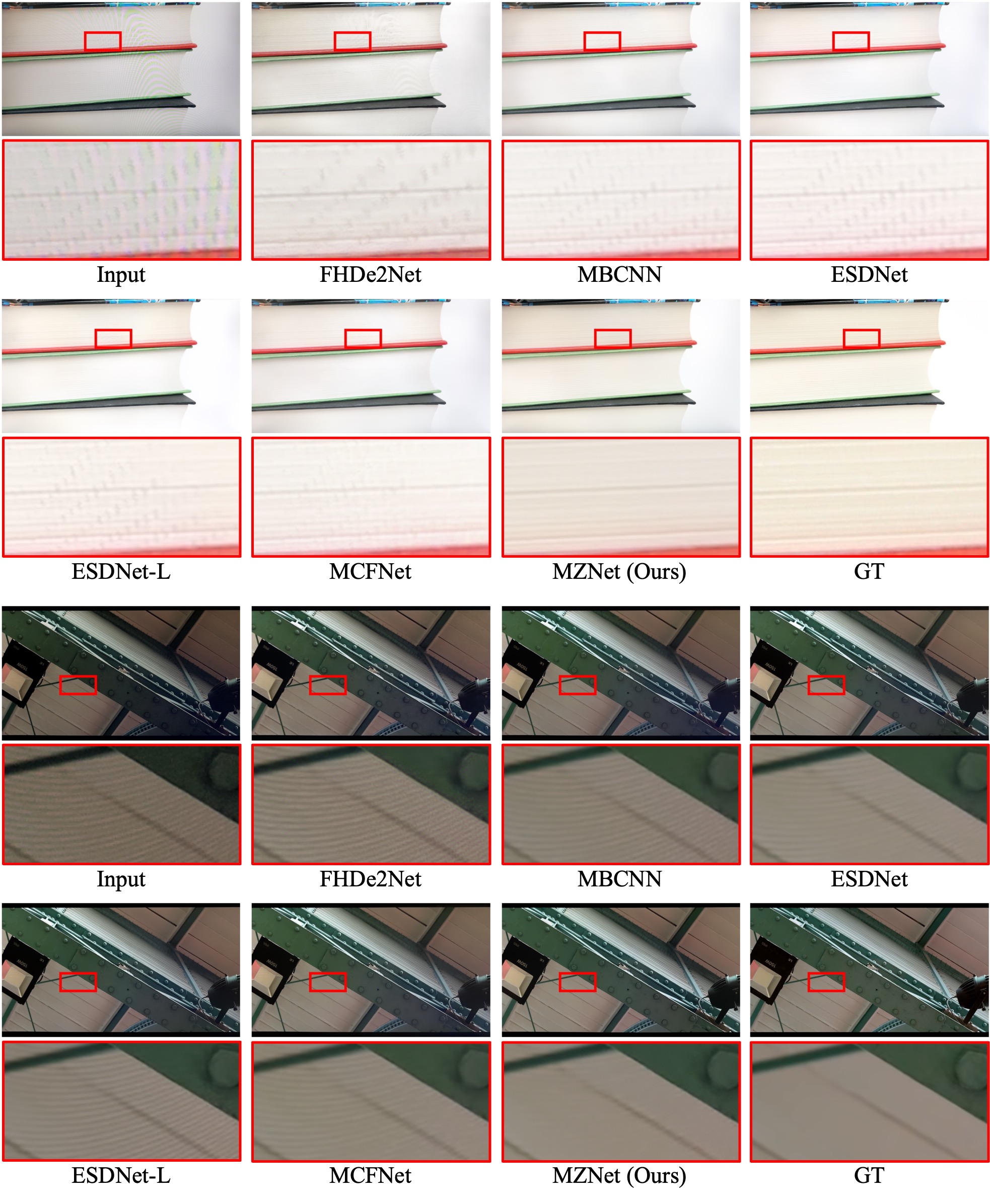}  
    \caption{Qualitative comparisons with state-of-the-art methods on the FHDMi dataset.}
    \label{fig:supple_fhdmi}
\end{figure*}

We present additional qualitative comparisons on the UHDM (\cref{fig:supple_uhdm,fig:supple_uhdm_2,fig:supple_uhdm_3,fig:supple_uhdm_4,fig:supple_uhdm_5}) and FHDMi (\cref{fig:supple_fhdmi}) datasets, with a particular focus on the UHDM, which consists of 4K images containing moir\'e patterns in various shapes and scales. 
Our method demonstrates superior performance in removing moir\'e artifacts, especially large-scale colorful patterns, compared to existing approaches. This validates that our model is well-designed for effective demoir\'eing.

\section{Limitations and Future work}
The real-world images were collected from actual display manufacturing processes under realistic operating conditions. However, MZNet has not yet been integrated into a complete industrial inspection workflow. In particular, we have not evaluated its effect on downstream process metrics such as defect detection accuracy, inspection success rate. A full quantitative assessment in a production environment is left for future work. 

In the current design, the kernel size $K$ of Multi-Shape Large Kernel Convolution Block (MSLKB) is manually adjusted based on the resolution of the training dataset to ensure sufficient receptive field coverage. While this approach allows for effective handling of various resolutions, it also requires dataset-specific tuning, potentially limiting the scalability and generalization of MZNet to unseen datasets or degradation scenarios. 

Furthermore, the architecture of MZNet is specifically tailored for moiré pattern removal, and its applicability to other types of degradations, such as JPEG artifacts, blur, and noise, has yet to be validated. Extending MZNet toward a more task-agnostic image restoration framework remains an important direction for future work.

Moreover, the potential of emerging architectures such as Mamba~\cite{gu2023mamba} for moiré removal has not yet been explored in this work, and we believe it presents a promising direction for future research.
\section{Broader Impact}

MZNet is designed to effectively remove moiré patterns. Beyond demonstrating strong performance on public datasets, MZNet is validated in real-world industrial scenarios, such as smartphone panel defect inspection, as detailed in main paper. This capability improves display inspection efficiency and reliability by removing the need for defocusing while ensuring high image quality for accurate defect detection. These improvements can positively impact industrial quality control by enabling precise defect detection, and supporting more sustainable and efficient workflows.
However, as with other image restoration methods, MZNet carries risks of misuse. While MZNet does not generate or manipulate images, its restoration process may unintentionally suppress subtle brightness variations or fine structural details that are important for detecting certain types of defects. To address this, we recommend using it within well-defined workflows, supported by clear guidelines and awareness of its limitations.



\end{document}